\documentclass[11pt]{article}

\usepackage[final]{acl}

\usepackage{times}
\usepackage{latexsym}
\usepackage[T1]{fontenc}
\usepackage[utf8]{inputenc}
\usepackage{microtype}
\usepackage{inconsolata}
\usepackage{graphicx}
\usepackage{booktabs}
\usepackage{enumitem}
\usepackage{url}

\title{GOD: Govern, Observe, and Direct ---\\
A Real-Time Control Room for Agent Societies}

\author{
  Yige Luo \and Ran Guan \\
  2012 Laboratories, Huawei \\
  \texttt{yl732@cantab.ac.uk} \quad \texttt{guanran@huawei.com} \\
}

\begin{document}
\maketitle

\begin{abstract}
Generative-agent systems are easier to start than to inspect.
A run can contain many agents, locations, messages, commands, and model calls, yet the operator often gets either a finished replay or raw logs.
That makes it hard to ask why an agent moved, test a small intervention, or package a run for another researcher.
GOD is a local-first control room for agent societies.
From the same browser workflow, an operator can issue targeted questions or interventions and inspect the resulting replay state.
The system combines a setup wizard, Agent Studio, Map Studio, a spatial replay interface, \textsc{Ask} and \textsc{Intervene} commands, and portable experiment, map, and agent packs.
Its technical contribution is the command and artifact loop: live controls and replay evidence share the same operator command model, while package contracts separate scenario, map, and profile data from local runtime state.
The public release includes hosted Smallville-style and PKU replays, the open-source repository, and downloadable packs.
We evaluate this path on 15 completed run slots. Across the 14 intervention runs, 78 of 84 target-agent checks recorded the commanded destination, and 169 of 182 state answers matched a saved location or action string.
\end{abstract}

\section{Introduction}
\label{sec:introduction}

Generative-agent systems can now populate simulated towns with language-model agents, but operating a run remains awkward.
An experiment may expose configuration files, a live process, logs, and a replay through separate interfaces.
Connecting a human command to the state written after that command then requires manual bookkeeping.
Generative Agents established the town pattern: agents retrieve stored experiences, form reflections, and plan behavior in a shared spatial environment \citep{park2023generativeagents}.
Other systems provide large-scale social simulation \citep{piao2025agentsociety}, social-interaction evaluation \citep{zhou2023sotopia}, and programmable multi-agent conversations \citep{wu2023autogen}.
GOD contributes an operator and artifact workflow that connects the bundled AgentSociety simulator with the JiuwenClaw runtime.

GOD is a local-first control room for agent societies.
The operator chooses or authors a scenario, inspects the current world, issues a targeted command, advances the run, and preserves the resulting state.
Agents remain visible on the map throughout this loop.
Every \textsc{Ask} and \textsc{Intervene} action is also recorded next to the corresponding replay step.

The intended users are NLP researchers, educators, designers, and agent-system builders who need a run that can be inspected without credentials and edited locally.
The hosted replay records what happened; the downloadable packs record the scenario, map, and agent profiles from which another run can be made.
These artifacts are linked in the interface but have different contracts.

We make three contributions:

\begin{enumerate}[leftmargin=*]
  \item \textbf{An operator command loop}: a browser control room connects mapped state, temporal controls, targeted questions, and next-step interventions.
  \item \textbf{A shared command record}: live controls and replay views use the same command schema, so questions and interventions remain attached to the state transitions they precede.
  \item \textbf{Separate artifact contracts}: runnable experiment, map, and agent packs are kept apart from replay history, credentials, and machine-local runtime state.
\end{enumerate}

\section{Related Work}
\label{sec:related}

GOD builds on systems for multi-agent conversation, simulation, and evaluation.
\begin{table*}[t]
  \centering
  \footnotesize
  \setlength{\tabcolsep}{4pt}
  \begin{tabular}{lccccc}
    \toprule
    \textbf{System} & \textbf{Spatial replay} & \textbf{Targeted ask} & \textbf{Live intervention} & \textbf{Browser/no-code} & \textbf{Portable artifacts} \\
    \midrule
    Generative Agents & R & R & P & NR & P \\
    AgentSociety & R & R & R & P & P \\
    AutoGen Studio & NR & P & P & R & R \\
    AgentScope & NR & R & P & R & R \\
    GOD & R & R & R & R & R \\
    \bottomrule
  \end{tabular}
  \caption{Feature-level comparison based on the cited primary papers and released GOD artifacts. R denotes an explicitly reported capability, P a related but non-equivalent affordance, and NR that the capability was not reported in the cited source; NR is not evidence of absence. Portable artifacts may be sessions, workflow configurations, replays, or scenario packages.}
  \label{tab:system-affordances}
\end{table*}

Table~\ref{tab:system-affordances} compares publicly described system affordances rather than agent quality or benchmark scores.
The evaluation does not claim a human-subject user study or a social-validity benchmark, and it does not claim that the agents are socially correct.

CAMEL and AutoGen organize interactions among language-model agents \citep{li2023camel,wu2023autogen}.
Generative Agents provides a persistent spatial town and natural-language interaction, while Concordia supplies a framework for grounded generative agent-based models \citep{park2023generativeagents,vezhnevets2023concordia}.
SOTOPIA evaluates social interaction, and AgentSociety combines large-scale simulation with surveys, interviews, and interventions \citep{zhou2023sotopia,piao2025agentsociety}.
GOD reuses these agent capabilities and focuses on a spatial operator and artifact workflow.

Developer tools also place interfaces around agent runtimes.
AutoGen Studio supports no-code workflow construction, debugging, and evaluation, while AgentScope provides development, monitoring, and deployment interfaces for multi-agent applications \citep{dibia2024autogenstudio,gao2024agentscope}.
GOD records live commands alongside spatial replay state and exports the scenario data separately.
Direct quantitative comparison would require adapters because these systems expose different command, state, and artifact schemas.

The released repository integrates trimmed AgentSociety and JiuwenClaw v0.1.11 subtrees, and its The Ville map assets are derived from the Generative Agents project \citep{piao2025agentsociety,openjiuwen2026jiuwenclaw,park2023generativeagents,zhang2026swarmskillsportableselfevolving}.
GOD does not introduce a new policy, planner, memory architecture, or base simulator.
It contributes the integration layer that maps browser commands to execution and replay records, the authoring views around that layer, and the import/export contracts for portable packs.

\section{System Overview}
\label{sec:system}

GOD uses a React/Vite control room and a FastAPI backend around the bundled AgentSociety simulator and JiuwenClaw agent runtime.
The backend connects browser commands to execution, replay storage, and pack import/export.
The operator supplies the language-model endpoint used during local execution.

\subsection{Design Goals}

GOD is designed around three practical constraints.
Operator actions should correspond to explicit system operations, such as a backend endpoint call, replay command entry, or package export.
Shared artifacts should be useful without exposing private runtime state.
The same scenario should support both a hosted browser replay for quick inspection and a local path for editing and reruns.

\begin{figure*}[t]
  \centering
  \includegraphics[width=0.96\linewidth]{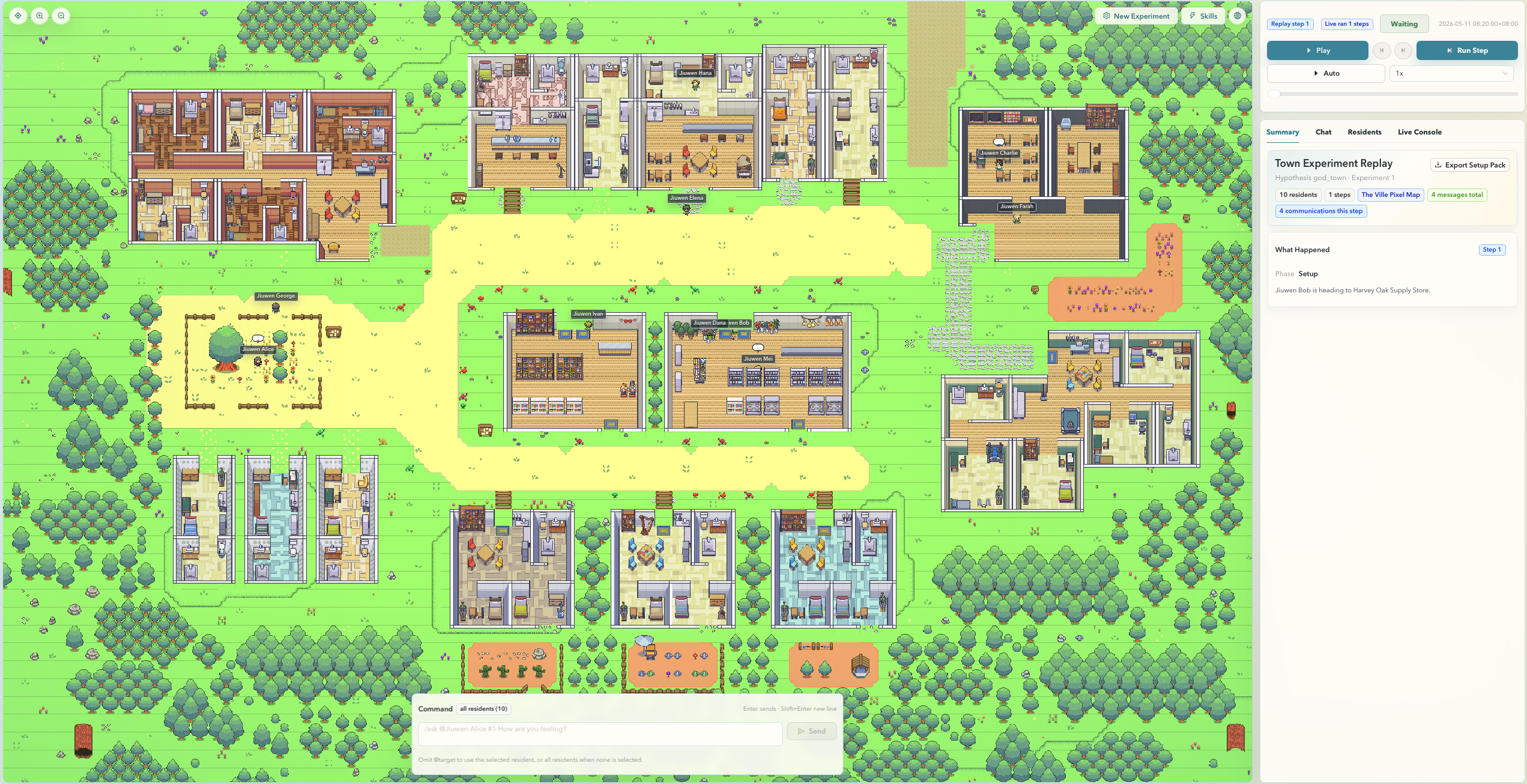}
  \caption{GOD control room for The Ville. The spatial state, replay and live-step controls, run summary, and shared command input remain visible in one browser view.}
  \label{fig:control-room}
\end{figure*}

\subsection{Operator Control Room}

The control room is the main interface for live and replayed runs.
Figure~\ref{fig:control-room} shows the full control-room layout: the world state remains visible beside temporal controls, run status, resident information, and the command input.
It exposes temporal controls for pausing, stepping, scrubbing, and replaying the simulation.
It also exposes two language-based controls:

\begin{itemize}[leftmargin=*]
  \item \textbf{Ask}: send a question to one agent, a group, or the full town.
  \item \textbf{Intervene}: inject an instruction into the next step so agents can react during execution.
\end{itemize}

Both commands are phrased in natural language.
For example, an operator can ask why a resident is in the park, ask for their view of an event or another agent, or announce that a nearby volcano erupted, and then inspect how agents react in the next replay frames.

\subsection{Setup Wizard}

The setup wizard collects model settings, scenario selection or creation, agent review, and launch in four steps.
It avoids hand-editing environment files for the standard local path.
Agent Studio edits resident profiles, while Map Studio edits map packages, locations, and collision constraints before local publication.
The same interface can select the bundled The Ville scenario, choose another experiment, open an imported pack, or publish a local experiment after editing agents and steps.
Appendix~\ref{app:authoring-views} shows the setup wizard, Map Studio, and Agent Studio views.

\begin{figure}[t]
  \centering
  \includegraphics[width=\linewidth]{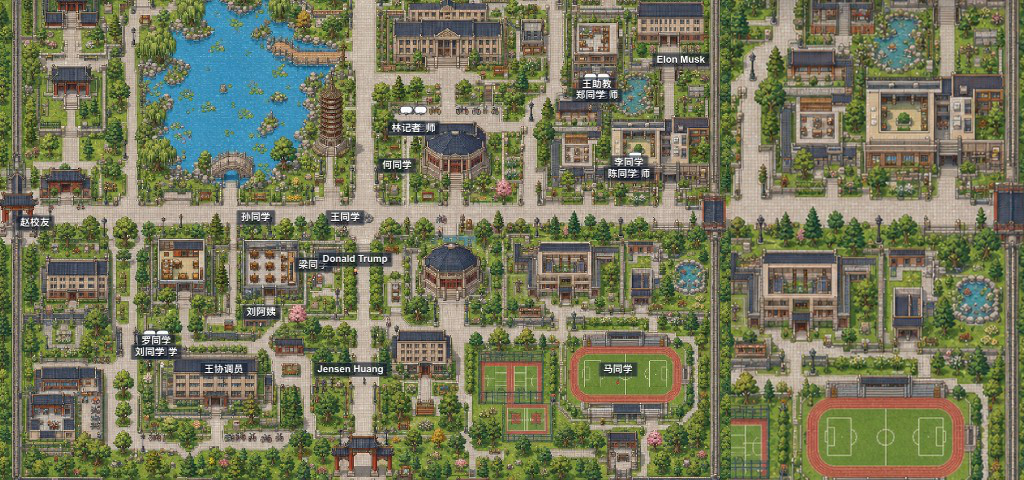}
  \caption{PKU campus map used by the 22-agent benchmark and public replay. Labels identify agents at their recorded positions in this frame.}
  \label{fig:map-examples}
\end{figure}

\subsection{Packs and Replays}

GOD separates experiments, maps, agents, replay stores, and runtime state.
Experiment packs define playable setups, map packs define geometry and assets, agent packs define resident profiles, and browser replays record completed runs.
Figure~\ref{fig:map-examples} shows the PKU map as rendered in a replay, while Figures~\ref{fig:map-studio} and \ref{fig:agent-studio} in the appendix show the local map and agent authoring tools.
Public packs include scenario, map, and profile data; they exclude API keys, local configuration, runtime snapshots, and replay databases.

A browser replay is a static record with a manifest, timeline, map metadata, character assets, resident profiles, and an operator command log.
An installable experiment pack is a runnable seed with scenario context, selected map, initial residents, initial locations, enabled skills, and step schedule.
This lets a reader inspect what happened without installation, then download packs to modify or reproduce the setup.

\subsection{Agent Runtime and Interaction Loop}

Each simulation step begins with an observation from the map environment.
For a resident, that observation includes the current location, tile position, nearby agents, recent messages, available map interactions, and the latest public event.
The agent prompt also receives the resident profile, shared world context, mounted skill IDs, and any pending operator interventions.
The profile contains identity, routine, relationships, needs, worries, quirks, recent history, and skill cues.

The JiuwenClaw agent runtime then chooses one mounted executable skill from the available catalog.
The catalog entry gives the skill description, declared effects, argument schema, and trigger examples.
The selected skill runs in the agent workspace and returns a structured result.
Those results can move an agent, run a location-scoped interaction, send a direct message to a nearby agent, send a public group message, update visible action/status/emotion fields, or record memory effects.
The environment applies these effects through explicit tools such as pathfinding, interaction execution, mailbox delivery, group broadcast, and state update.
After all agents finish the step, Pixel Town writes the replay frame: agent locations, movement segments, actions, emotions, latest event, and communications.

\subsection{Ask and Intervene}

\textsc{Ask} is implemented as a read-only interview over the same live state.
For a targeted resident, the backend calls the agent's external-question path with the current simulation time and asks for a first-person answer grounded in the profile, current context, recent questions, and session state.
For a society-level question, the call goes through the simulation router.
In both cases, the command does not modify the environment.

\textsc{Intervene} changes what the next step can see or do.
Movement-style commands, such as asking one or more agents to gather at a named location, call the map environment's pathfinder directly and expose the resulting path length.
World-event commands call the environment event publisher, which can set the current phase, add the event to later observations, and broadcast it to the group mailbox.
Other targeted instructions are stored as pending interventions on the selected agents and are inserted into their next step prompt.
Each command is stored as a record with the same schema in both modes: target, prompt, result, simulation time, and step.
Both the live control room and the replay log render this record, so each command is shown beside its recorded simulation step.

\subsection{Runtime Boundary}

The implementation keeps authoring and execution local.
The browser calls a FastAPI live-session API for run, step, ask, intervene, auto-run, pause, and stop operations; replay browsing uses separate read-only endpoints for metadata, datasets, map assets, sprites, and timeline frames.
Model endpoints, API keys, ports, generated sprites, and transient replay stores stay on the operator's machine.
Hosted replays require no credentials; authoring, live commands, and new simulation steps require a local run.

\section{Demonstration Plan}
\label{sec:demo}

The demonstration uses the fictional GOD Town scenario, whose setting and visual assets are adapted from Generative Agents \citep{park2023generativeagents}.
The town has 10 residents, 10 semantic locations, and 65 location-scoped interactions.
Each resident has a profile with identity, routine, relationships, needs, worries, quirks, recent history, and mounted skill cues.
The run starts on an ordinary weekday morning and advances in 30-minute in-world ticks.

The local demo follows one operator cycle and then opens the authoring workflow.
The operator inspects the replay summary, messages, residents, and one profile before distinguishing stored-frame playback from a live step.
A read-only question to Alice records the current answer without changing the world.
A group intervention asks all 10 residents to gather at the library; automatic live steps advance the world while the command record remains beside the new map frames.
The operator then inspects a registered skill and uses the setup workflow to generate a map draft, surface route-validation warnings, and create an editable experiment draft.
The walkthrough stops at launch preflight and returns to the current run; pack export remains separate from replay history and is not executed in the video.

The same scenario is available as a hosted read-only replay at \url{https://xiaoluolyg.github.io/GOD/replays/god-town/}.
Editing maps, changing packs, running new steps, and comparing variants require local execution.

\subsection{Audience and Use Cases}

GOD is intended for controlled scenario comparisons, classroom walkthroughs, and interactive-world design.
An operator can hold the map and profiles fixed while changing an event, its recipients, its timing, or a destination.
The resulting runs are simulation artifacts for inspection and teaching, not substitutes for studies of human behavior.

\section{Evaluation}
\label{sec:evaluation}

GOD integrates an existing simulator and agent runtime, so we evaluate its command-to-artifact path, not agent intelligence.
We ask three questions: (Q1) do operator commands leave measurable state and command records, (Q2) can saved interviews be checked against replay state and event boundaries, and (Q3) do the artifacts support controlled reruns and package validation?

\subsection{End-to-End Protocol}

The final scored set contains 15 completed run slots from GOD's live backend: one no-event baseline and 14 intervention runs from 10 scenario templates, four of which were run twice.
Every run used the same PKU map, 22 profiles, and initial locations.
All scored runs used Qwen-Plus through the DashScope endpoint.
We did not override the provider's default temperature, and the model configuration was held fixed across scenarios.
The scenarios cover public warnings, institutional notices, a false rumor, and controlled changes to notification target, notification time, or movement destination.
The runner calls the same \textsc{Ask}, \textsc{Intervene}, and \textsc{Run Step} endpoints as the browser, then reads the resulting SQLite replays and command transcripts.
Appendix~\ref{app:benchmark-details} gives the full protocol.

\subsection{Measures}

\textbf{Event routing} checks the command transcript to distinguish event or targeted-instruction delivery from the movement path.
\textbf{Event-specific trace@2} counts agent-run pairs whose replay fields in either of the first two post-event frames contain a strong scenario term, or whose event-belief answer contains one without an explicit denial.
All 22 agents contribute replay fields, but only the five interviewed agents can also satisfy the answer branch; the staff-only variant still uses all 22 agents as its denominator although only six received the instruction.
\textbf{Target destination recorded@1} checks whether a targeted agent has the commanded destination as its current or target location after one movement step.
The corresponding non-target check can include an agent who was already at that location, so it is not a causal side-effect measure.

The interview measures are deterministic string checks.
\textbf{Replay-state match} looks for a location alias or action substring from the nearest replay frame.
\textbf{Event-boundary match} checks that pre-event-awareness answers deny or omit strong event terms, and that post-event-belief and post-movement why-location answers contain one without a denial.
We also count mentions from a fixed list of unsupported locations, post-event answers without a role-related profile term, and strong event terms before the intervention.
These labels do not establish semantic grounding, hallucination, or persona quality.
The results describe this Qwen-Plus configuration rather than model-independent agent behavior.
The denominators are repeated checks over fixed simulated profiles, not independent subjects: the same 22 profiles recur across scenarios, and interviews reuse five pre/post agents plus three movement targets.

\begin{table}[t]
  \centering
  \scriptsize
  \begin{tabular}{lr}
    \toprule
    \textbf{Measure} & \textbf{Value} \\
    \midrule
    Completed planned run slots & 15 / 15 \\
    Event commands avoiding movement routing & 14 / 14 \\
    Event-specific trace@2 & 212 / 308 agent-runs \\
    Target destination recorded@1 & 78 / 84 targets \\
    Non-target at commanded destination & 23 / 224 non-targets \\
    Replay-state string match & 169 / 182 answers \\
    Event-boundary string match & 144 / 182 answers \\
    Unsupported-location mention & 0 / 392 answers \\
    Post-answer role-anchor miss & 9 / 252 answers \\
    Pre-event event-term leakage & 0 / 140 answers \\
    Pairwise repeat location JSD & 0.011 \\
    \bottomrule
  \end{tabular}
  \caption{Command-to-artifact results. Counts pool the 14 completed intervention runs; four scenario templates contribute two runs. JSD is the mean over four pairs of final agent-location distributions.}
  \label{tab:operator-benchmark}
\end{table}

\subsection{Results}

The first pass exposed a routing collision: the parser trigger \emph{dao} (``to/arrive'') also occurred inside the Chinese compounds \emph{shou-dao} (``receive'') and \emph{dao-fang} (``visit'').
We added a regression test and reran the six affected runs from fresh directories.
After the fix, all 14 event commands reached the event or targeted-instruction path.
This 14/14 result is therefore a regression check on the corrected command path, not a held-out estimate of routing generalization.
One discarded attempt of the delayed-notice run exceeded the 360-second request limit during a pre-event step; its fresh retry completed, and only completed-run markers enter Table~\ref{tab:operator-benchmark}.

Movement targets were recorded in 78 of 84 target-agent checks.
All six misses came from the gymnasium variant, where pathfinding reported the destination as unreachable.
The 23 non-target matches are reported separately because they include agents already at the commanded destination.

Strong event terms appeared in 212 of 308 agent-run evidence windows.
The staff-only notice left no such term in its two-step window even though its command response records targeted acceptance for the six selected agents.
This distinction is why routing and trace evidence are separate measures.

For Q2, 169 of 182 state answers matched a saved location alias or action substring, and 144 of 182 event-boundary answers passed the stated term rule.
The latter combines 70/70 pre-event absence checks with 74/112 post-intervention presence checks; the separate 0/140 leakage audit scans all ten pre-event answers per run.
The zero unsupported-location count applies only to the seven listed locations.
For Q3, the mean pairwise final-location JSD across the four repeated scenario pairs is 0.011.
Three pairs had JSD 0, while the diplomatic-visit pair had 0.045.
This permits a compact replay comparison; four pairs do not establish deterministic behavior, stability, or a timing effect.

Table~\ref{tab:operator-benchmark} uses pooled run-level counts.
The released result file also labels its separate scenario-macro summary: for example, trace is 212/308 (68.8\%) in the pooled table and 72.7\% when repeats are averaged within a scenario first.

\subsection{Comparison and Artifact Checks}

Because the runner assumes GOD's command, replay, profile, and pack schemas, scores for other systems would require adapters to the same evidence fields.
No cross-system scores are reported.
Table~\ref{tab:system-affordances} instead compares capabilities reported in the cited system descriptions.

The selected backend suite passed 82 tests covering replay export, package import/export, live endpoints, operator commands, and setup routing.
The static build also passed replay and package validation.
Appendix~\ref{app:artifact-checks} lists the checked artifacts and the private state excluded from public packs.

\section{Conclusion}
\label{sec:conclusion}

GOD connects local scenario setup and live operator commands to replay inspection and portable packs for generative-agent societies.
Its contribution is the operator and artifact workflow over the integrated simulator and agent runtime, rather than a new town simulation or agent policy.
The selected backend suite passed 82 tests covering the released replay and package paths.
The live benchmark used one model configuration and four repeated scenario pairs; operator usability was not evaluated.

\paragraph{Availability and licensing.}
GOD is released under Apache-2.0 at \url{https://github.com/XiaoLuoLYG/GOD}.
The public site is available at \url{https://xiaoluolyg.github.io/GOD/}, with hosted replays and downloadable experiment, map, and agent packs.
Hosted replays require no credentials, whereas new runs require a local model endpoint; third-party subtrees and derived visual assets retain their upstream licenses and attributions.

\section{Limitations}

We tested one model configuration with repeated synthetic profiles.
Target destination recorded@1 checks the recorded state after one movement step rather than completed trajectory execution or arrival.
The string checks measure command and replay consistency, not semantic grounding or operator usability.
Results may differ with other models, scenarios, and users.

\section{Ethics and Broader Impact}

Because GOD can simulate real places, public situations, and named people, readers may mistake generated behavior for evidence about actual people or institutions.
The main demo therefore uses a fictional town and synthetic residents and labels their behavior as simulated.
Scenarios involving real institutions, public events, or named people use synthetic profiles and serve only as packaging and replay stress tests.
We describe these scenarios as stylized artifacts with authored assumptions and keep the qualitative case study on the Smallville-style town.
Prompts sent to a hosted model remain subject to that provider's data policy.

Reviewers can inspect public replay data and packs without access to local runtime state: the packs include scenario, asset, and profile data but exclude API keys, logs, private replay databases, and machine-specific configuration.

\bibliography{references}

@misc{park2023generativeagents,
  title         = {Generative Agents: Interactive Simulacra of Human Behavior},
  author        = {Park, Joon Sung and O'Brien, Joseph C. and Cai, Carrie J. and Morris, Meredith Ringel and Liang, Percy and Bernstein, Michael S.},
  year          = {2023},
  eprint        = {2304.03442},
  archivePrefix = {arXiv},
  primaryClass  = {cs.HC},
  doi           = {10.48550/arXiv.2304.03442},
  url           = {https://arxiv.org/abs/2304.03442}
}

@inproceedings{li2023camel,
  title     = {CAMEL: Communicative Agents for "Mind" Exploration of Large Language Model Society},
  author    = {Li, Guohao and Hammoud, Hasan Abed Al Kader and Itani, Hani and Khizbullin, Dmitrii and Ghanem, Bernard},
  booktitle = {Advances in Neural Information Processing Systems},
  year      = {2023},
  url       = {https://proceedings.neurips.cc/paper_files/paper/2023/hash/a3621ee907def47c1b952ade25c67698-Abstract-Conference.html}
}

@misc{vezhnevets2023concordia,
  title         = {Generative Agent-Based Modeling with Actions Grounded in Physical, Social, or Digital Space Using Concordia},
  author        = {Vezhnevets, Alexander Sasha and Agapiou, John P. and Aharon, Avia and Ziv, Ron and Matyas, Jayd and Duenez-Guzman, Edgar A. and Cunningham, William A. and Osindero, Simon and Karmon, Danny and Leibo, Joel Z.},
  year          = {2023},
  eprint        = {2312.03664},
  archivePrefix = {arXiv},
  primaryClass  = {cs.AI},
  doi           = {10.48550/arXiv.2312.03664},
  url           = {https://arxiv.org/abs/2312.03664}
}

@article{piao2025agentsociety,
  title         = {AgentSociety: Large-Scale Simulation of LLM-Driven Generative Agents Advances Understanding of Human Behaviors and Society},
  author        = {Piao, Jinghua and Yan, Yuwei and Zhang, Jun and Li, Nian and Yan, Junbo and Lan, Xiaochong and Lu, Zhihong and Zheng, Zhiheng and Wang, Jing Yi and Zhou, Di and Gao, Chen and Xu, Fengli and Zhang, Fang and Rong, Ke and Su, Jun and Li, Yong},
  journal       = {arXiv preprint arXiv:2502.08691},
  year          = {2025},
  eprint        = {2502.08691},
  archivePrefix = {arXiv},
  primaryClass  = {cs.SI},
  doi           = {10.48550/arXiv.2502.08691},
  url           = {https://arxiv.org/abs/2502.08691}
}

@article{zhou2023sotopia,
  title         = {SOTOPIA: Interactive Evaluation for Social Intelligence in Language Agents},
  author        = {Zhou, Xuhui and Zhu, Hao and Mathur, Leena and Zhang, Ruohong and Yu, Haofei and Qi, Zhengyang and Morency, Louis-Philippe and Bisk, Yonatan and Fried, Daniel and Neubig, Graham and Sap, Maarten},
  journal       = {arXiv preprint arXiv:2310.11667},
  year          = {2023},
  eprint        = {2310.11667},
  archivePrefix = {arXiv},
  primaryClass  = {cs.AI},
  doi           = {10.48550/arXiv.2310.11667},
  url           = {https://arxiv.org/abs/2310.11667}
}

@article{wu2023autogen,
  title         = {AutoGen: Enabling Next-Gen LLM Applications via Multi-Agent Conversation},
  author        = {Wu, Qingyun and Bansal, Gagan and Zhang, Jieyu and Wu, Yiran and Li, Beibin and Zhu, Erkang and Jiang, Li and Zhang, Xiaoyun and Zhang, Shaokun and Liu, Jiale and Awadallah, Ahmed Hassan and White, Ryen W. and Burger, Doug and Wang, Chi},
  journal       = {arXiv preprint arXiv:2308.08155},
  year          = {2023},
  eprint        = {2308.08155},
  archivePrefix = {arXiv},
  primaryClass  = {cs.AI},
  doi           = {10.48550/arXiv.2308.08155},
  url           = {https://arxiv.org/abs/2308.08155}
}

@misc{dibia2024autogenstudio,
  title         = {{AutoGen Studio}: A No-Code Developer Tool for Building and Debugging Multi-Agent Systems},
  author        = {Dibia, Victor and Chen, Jingya and Bansal, Gagan and Syed, Suff and Fourney, Adam and Zhu, Erkang and Wang, Chi and Amershi, Saleema},
  year          = {2024},
  eprint        = {2408.15247},
  archivePrefix = {arXiv},
  primaryClass  = {cs.SE},
  doi           = {10.48550/arXiv.2408.15247},
  url           = {https://arxiv.org/abs/2408.15247}
}

@misc{gao2024agentscope,
  title         = {{AgentScope}: A Flexible yet Robust Multi-Agent Platform},
  author        = {Gao, Dawei and Li, Zitao and Pan, Xuchen and Kuang, Weirui and Ma, Zhijian and Qian, Bingchen and Wei, Fei and Zhang, Wenhao and Xie, Yuexiang and Chen, Daoyuan and Yao, Liuyi and Peng, Hongyi and Zhang, Zeyu and Zhu, Lin and Cheng, Chen and Shi, Hongzhu and Li, Yaliang and Ding, Bolin and Zhou, Jingren},
  year          = {2024},
  eprint        = {2402.14034},
  archivePrefix = {arXiv},
  primaryClass  = {cs.MA},
  doi           = {10.48550/arXiv.2402.14034},
  url           = {https://arxiv.org/abs/2402.14034}
}

@misc{openjiuwen2026jiuwenclaw,
  title  = {{JiuwenClaw}},
  author = {{openJiuwen}},
  year   = {2026},
  note   = {Software release, version 0.1.11},
  url    = {https://pypi.org/project/jiuwenclaw/0.1.11/}
}

@misc{zhang2026swarmskillsportableselfevolving,
  title         = {Swarm Skills: A Portable, Self-Evolving Multi-Agent System Specification for Coordination Engineering},
  author        = {Zhang, Xinyu and Dou, Zhicheng and Li, Deyang and Tao, Jianjun and Cheng, Shuo and Shi, Ruifeng and Liu, Fangchao and Hu, Enrui and Ding, Yangkai and Wang, Hongbo and Ye, Qi and Jin, Xuefeng and Zhao, Zhangchun},
  year          = {2026},
  eprint        = {2605.10052},
  archivePrefix = {arXiv},
  primaryClass  = {cs.CL},
  doi           = {10.48550/arXiv.2605.10052},
  url           = {https://arxiv.org/abs/2605.10052}
}

\clearpage
\appendix
\section{Authoring Views}
\label{app:authoring-views}

The views below show the setup wizard's model step and the local tools for editing map and agent packages before launch.

\begin{center}
  \centering
  \includegraphics[width=\linewidth]{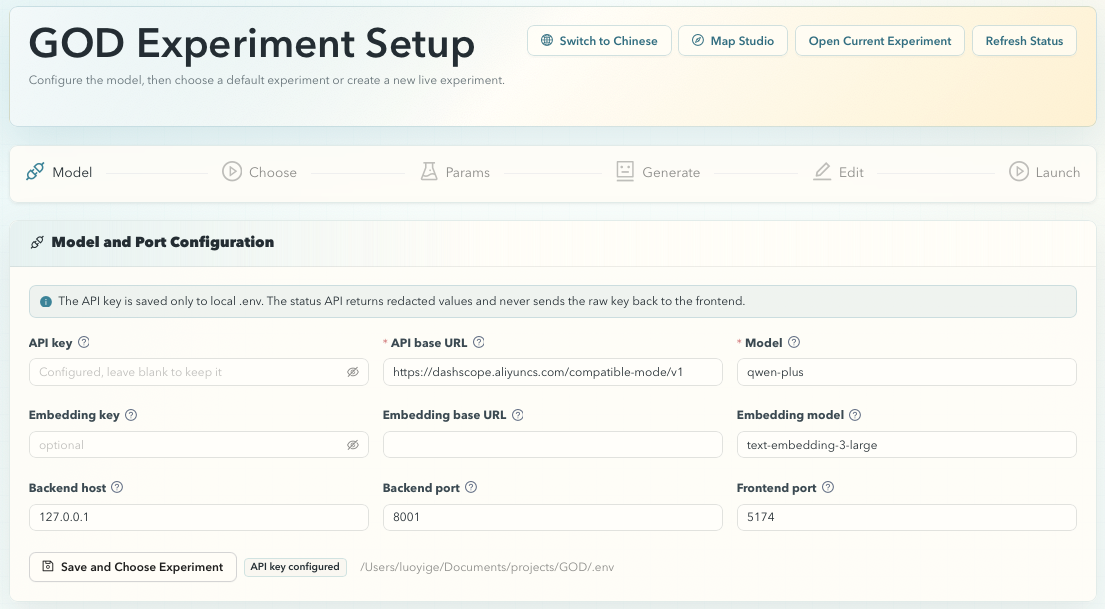}
  \captionof{figure}{Setup wizard for local model configuration. The first step records the model endpoint and local ports before the operator chooses a bundled experiment, imports a pack, or creates a new run.}
  \label{fig:setup-wizard}
\end{center}

\begin{center}
  \centering
  \includegraphics[width=\linewidth]{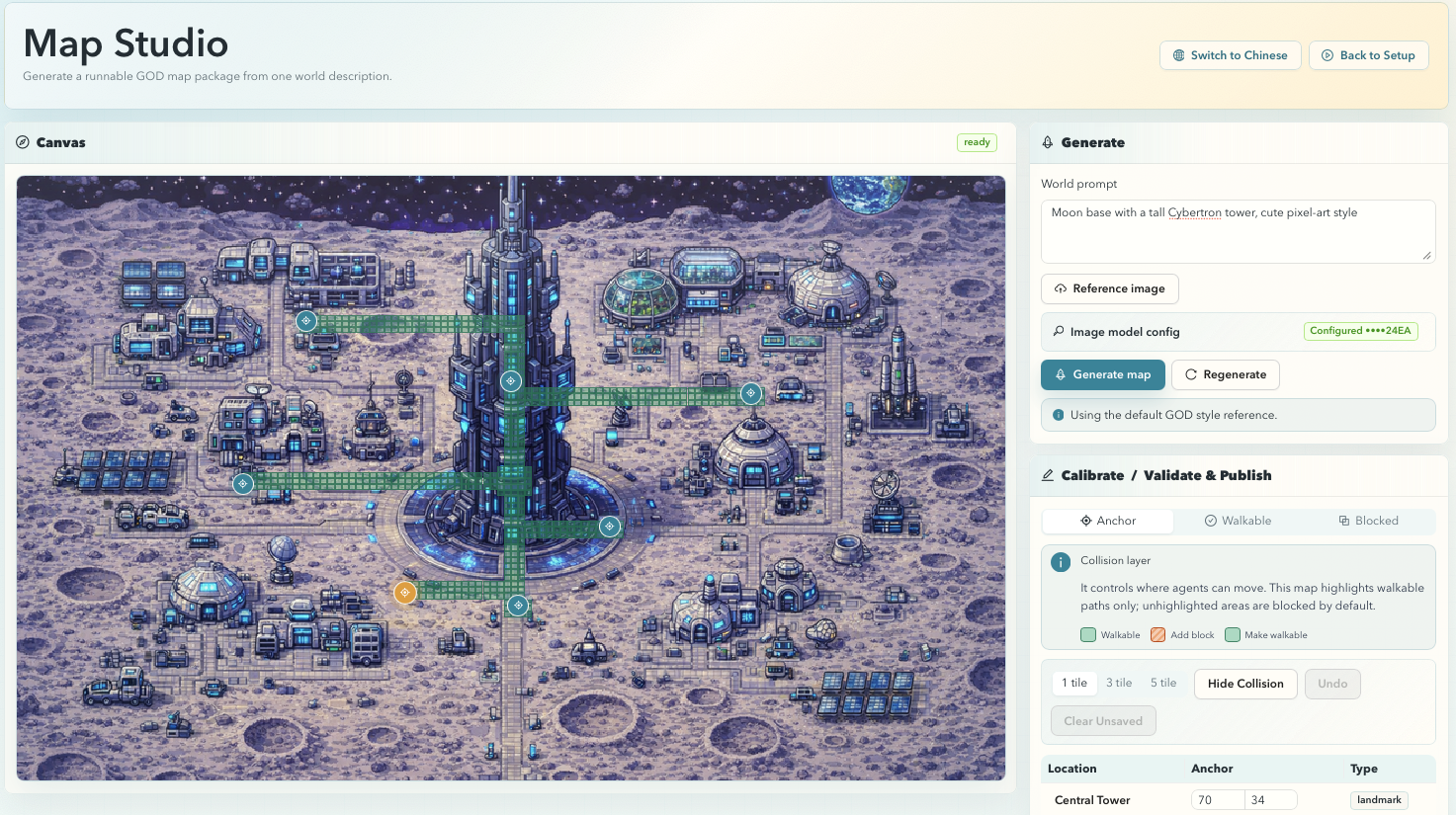}
  \captionof{figure}{Map Studio turns a generated or imported map into a runnable package by calibrating locations, walkable paths, and blocked regions.}
  \label{fig:map-studio}
\end{center}

\begin{center}
  \centering
  \includegraphics[height=0.5\linewidth]{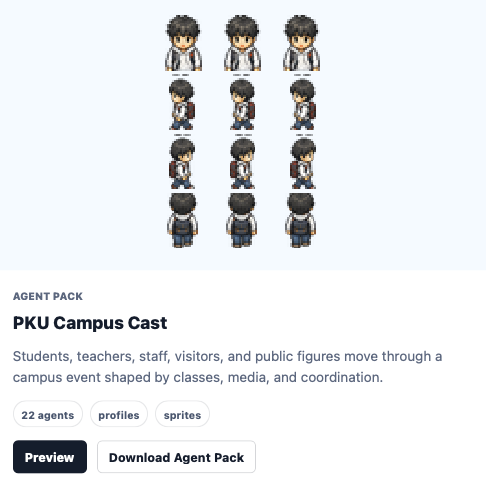}%
  \includegraphics[height=0.5\linewidth]{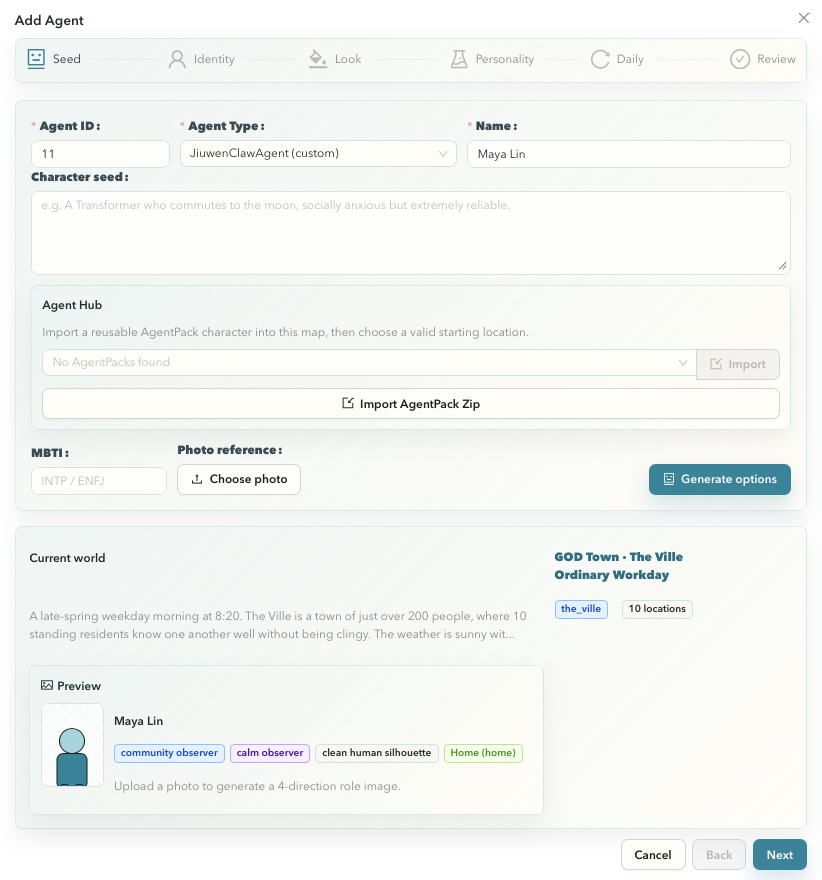}
  \captionof{figure}{PKU agent pack and Agent Studio examples. Agent packs distribute reusable profiles and sprites; Agent Studio lets an operator create or import a resident, edit identity and personality fields, and place the agent into the current world.}
  \label{fig:agent-studio}
\end{center}

\section{Benchmark and Artifact Details}
\label{app:benchmark-details}

The benchmark runner creates experiments under \texttt{hypothesis\_god\_full\_benchmark\_20260708}, calls the live \textsc{Ask}, \textsc{Intervene}, and \textsc{Run Step} endpoints used by the browser, and computes metrics from each run's SQLite replay and command transcript.
Runs are stored in separate scenario/repeat experiment directories, and only transcripts containing a completed-run marker enter scoring.

All runs use the same PKU campus map, the same 22 agent profiles, and the same initial locations.
There is one no-event baseline.
The event-type block includes a volcano warning, an earthquake warning, a school power outage, a diplomatic visit, a rumor that an event was canceled, a public lecture, and a traffic blockade.
Three controlled variants change one variable at a time: notification target (staff-only diplomatic notice), notification time (two ordinary steps before the diplomatic notice), and destination (public lecture moved to the gymnasium).
The repeat block runs four scenarios twice: volcano, earthquake, diplomatic visit, and rumor.

For each non-baseline run, the operator first asks five agents with different roles whether they know of a public event and where they are.
The roles are student, teaching assistant, librarian, reporter, and coordinator.
After injecting the event, the operator advances two steps and asks the same agents about their state, event belief, and one social relation.
A movement intervention then targets the scenario destination.
After one more step, the operator asks three target agents why they are in their current location.
The delayed-notice variant adds two ordinary pre-event steps before the intervention.
Thus event metrics are measured at @2, while target-destination recording is measured at @1.

The main table reports pooled numerators and denominators over the 14 completed intervention runs, so each repeated run remains visible in the sample size.
The released metrics file also reports scenario-level macro-averages, averaging repeats within a scenario before averaging the 10 intervention scenarios.
Event-specific trace coverage uses strong scenario terms in the first two post-event replay frames or in the event-belief interview at that boundary; explicit denials are excluded from the interview branch.
Replay-state match tests saved location aliases and action substrings against the nearest frame.
Event-boundary match checks pre-event-awareness, post-event-belief, and post-movement why-location answers.
Unsupported-location mention, role-anchor miss, and pre-event leakage are likewise lexical checks.
The repeat JSD is the mean of the four pairwise divergences for scenarios run twice; single-run scenarios do not enter that number.

\subsection{Artifact and Reproducibility Checks}
\label{app:artifact-checks}

The public site exposes two replay pages with 29 timeline frames, 10 and 22 agent entries, four completed operator commands, and eight release-backed replay downloads.
The pack library exposes 10 experiment packs, 10 map packs, and 10 agent packs, containing 141 profile entries, 104 location entries, and 173 interaction entries.
The validator checks manifest fields, timeline length, step files, command completion status, profile counts, map metadata, and release-backed download links.
It completed without errors on the current static build.

We also ran backend tests for public replay export, package import and export, experiment packs, map packs, agent packs, live experiment endpoints, operator commands, and setup wizard routing.
These tests exercise the boundary between portable data and local runtime state.
Experiment, map, and agent packs contain scenario, map, and profile data.
They exclude local logs, SQLite replay stores, runtime snapshots, API keys, model credentials, and machine-specific paths.
A release-boundary audit scanned 635 public-data files and 141 public agent runtime configs for private-state files, local paths, and secret-like assignments, and found no violations.

\end{document}